\documentclass[10pt,twocolumn,british,preprint]{IEEEtran}
\usepackage[T1]{fontenc}
\usepackage[latin9]{inputenc}
\usepackage{babel}
\usepackage{booktabs}
\usepackage{graphicx}
\usepackage[numbers]{natbib}
\usepackage[unicode=true]
 {hyperref}

\makeatletter

\providecommand{\tabularnewline}{\\}
\newcommand{\lyxdot}{.}

\makeatother

\begin{document}
\title{Whole-examination AI estimation of fetal biometrics from 20-week ultrasound
scans}
\author{\IEEEauthorblockN{Lorenzo Venturini}\IEEEauthorrefmark{1} , \IEEEauthorblockN{Samuel Budd}\IEEEauthorrefmark{1},
\IEEEauthorblockN{Alfonso Farruggia}\IEEEauthorrefmark{1}, \IEEEauthorblockN{Robert Wright}\IEEEauthorrefmark{1},
\IEEEauthorblockN{Jacqueline Matthew}\IEEEauthorrefmark{1}, \IEEEauthorblockN{Thomas G. Day}\IEEEauthorrefmark{1}\IEEEauthorrefmark{2},
\IEEEauthorblockN{Bernhard Kainz}\IEEEauthorrefmark{1}\IEEEauthorrefmark{3},
\IEEEauthorblockN{Reza Razavi}\IEEEauthorrefmark{1}\IEEEauthorrefmark{2},
\IEEEauthorblockN{Jo V. Hajnal}\IEEEauthorrefmark{1}\\\IEEEauthorrefmark{1}\IEEEauthorblockA{Department of Imaging Sciences, Faculty of Life Sciences and Medicine,
King's College London} \\\IEEEauthorrefmark{2}\IEEEauthorblockA{Guy's and St Thomas' NHS Foundation Trust, London}\\\IEEEauthorrefmark{3}\IEEEauthorblockA{Department of Computing, Faculty of Engineering, Imperial College
London, London}}
\maketitle
\begin{abstract}
The current approach to fetal anomaly screening is based on biometric
measurements derived from individually selected ultrasound images.
In this paper, we introduce a paradigm shift that attains human-level
performance in biometric measurement by aggregating automatically
extracted biometrics from every frame across an entire scan, with
no need for operator intervention. We use a convolutional neural network
to classify each frame of an ultrasound video recording. We then measure
fetal biometrics in every frame where appropriate anatomy is visible.
We use a Bayesian method to estimate the true value of each biometric
from a large number of measurements and probabilistically reject outliers.
We performed a retrospective experiment on 1457 recordings (comprising
48 million frames) of 20-week ultrasound scans, estimated fetal biometrics
in those scans and compared our estimates to the measurements sonographers
took during the scan. Our method achieves human-level performance
in estimating fetal biometrics and estimates well-calibrated credible
intervals in which the true biometric value is expected to lie.
\end{abstract}


\section{Introduction}

Ultrasound (US) imaging is routinely used in many countries during
pregnancy to screen for fetal abnormalities, often at \textasciitilde 18-22
weeks \citep{Salomon2019}. Procedures typically involve imaging of
standard planes of fetal anatomy and measurement of several fetal
biometrics. In the UK, requirements for screening examinations are
published in the Fetal Anatomy Screening Programme (FASP) standard
\citep{FASP}. 

Fetal biometrics are conventionally measured in single images: the
operator pauses the US image stream on a specified view, then manipulates
calipers to measure the anatomy. Some guidelines recommend that this
whole procedure is repeated up to three times to ensure reliability
\citep{Salomon2022}. Traditionally, these measurements have been
done manually. Manual measurement of fetal biometrics displays significant
expected-value bias \citep{Drukker2020}. Selection bias may also
influence the planes used for measurement. Furthermore, these manual
processes display significant inter-observer variability, ranging
from 4.9\% to 11.1\% depending on the measurement \citep{Sarris2012}.

Recent research has proposed methods for automating the detection
of standard anatomical planes in during live scanning live scanning
\citep{Baumgartner2017,Chen2015,Burgos2020}and biometric measurements
\citep{Sinclair2018}. US machine manufacturers are increasingly integrating
these tools in their equipment \citep{Yaqub2021}. However, fetal
biometry generally still follows existing workflows that rely on operator
selection of individual images. This does not take full advantage
of the real-time nature of US, where a video stream of tens of frames
per second is acquired. A fully automatic system could perform plane
classification on every frame of the US stream and biometric measurement
using all available data.

Fully automation would reduce human selection bias from plane selection
and expected-value bias during measurement, potentially improving
veracity, reproducibility and reducing operator dependence. Automating
some key operator tasks could also reduce cognitive load, allowing
the sonographer to focus more on the patient and on identifying signs
of anomalies.

Evaluating every frame in a fully automatic system generates a very
large number of measurements. A key challenge in making such a system
useful for clinical practice, therefore, is to generate a single estimate
from the vast resulting amount of information. There has been some
work in developing methods to calculate fetal biometry from a video
feed. P\l otka et al \citep{Plotka2021} propose a system to select
the best standard planes to extract biometry, and Lee et al \citep{Lee2023}
average convolutional neural network (CNN) outputs across a scan to
estimate gestational age. Matthew et al \citep{Matthew2022} have
proposed a system to automatically classify every frame in a US scan
and extract fetal biometrics, but an operator still needs to select
appropriate frames to report biometrics.

No work has, to our knowledge, sought to obtain an expected value
of the biometry and a credible interval by combining all frame measurements
while acknowledging potential outliers. This would represent a substantial
departure from current clinical practice, which is fully focused on
a selection and measurement of single images.

\subsection{Contributions}

We propose a real-time system that can identify standard planes and
estimate fetal biometrics achieving and reporting progressively reducing
uncertainties. The system is designed to seamlessly link into clinical
practice during 20-week US screening scanning. We build upon Sononet
\citep{Baumgartner2017} for standard plane detection, and pair it
with automatic biometric estimation per frame, aggregating these to
generate a progressive global estimate using a Bayesian framework.
When used during live scanning, this results in progressively more
reliable central estimates and credible intervals for each biometric.
The proposed method has the potential to improve clinical practice
by providing robust measurements free from expectation bias, while
reducing cognitive load on the sonographer.

\section{Methods}

\subsection{Datasets}

\begin{table*}
\begin{centering}
\begin{tabular}{ccccc}
\toprule 
Data fold & Train & Validation & Test & lowest p-value\tabularnewline
\midrule
\midrule 
Total scans & 4395 & 1457 & 1457 & N/A\tabularnewline
\midrule 
Abnormalities & 528 (12.0\%) & 155 (10.6\%) & 163 (11.2\%) & 0.360\tabularnewline
\midrule 
Live birth & 4363 (99.2\%) & 1442 (98.9\%) & 1450 (99.5\%) & 0.375\tabularnewline
\midrule 
Avg. GA & 20w 3d & 20w 3d & 20w 3d & 0.653\tabularnewline
\midrule 
Nonwhite & 1352 (30.8\%) & 458 (31.4\%) & 418 (28.7\%) & 0.166\tabularnewline
\midrule 
Avg. BMI (kg / m\textsuperscript{2}) & 24.7 & 24.9 & 24.9 & 0.448\tabularnewline
\bottomrule
\end{tabular}
\par\end{centering}
\medskip{}

\caption{Patient characteristics and metadata across the data folds used for
model training. None of the patient characteristics, including gestational
age (GA), maternal body mass index (BMI), or percentage of abnormal
scans were significantly different across the chosen folds.\label{tab:Patient-metadata}}
\end{table*}

The data for this work is taken from the iFIND project on fetal US,
which recruited mothers attending a routine 20-week anomaly screening
clinic in London, UK. 10000 volunteers gave consent to have their
full 20-week US examinations (gestational age 18-22 weeks) recorded\footnote{\href{http://www.ifindproject.com}{www.ifindproject.com}}.The
examinations were conducted on identical US machines (GE Voluson E8)
by 145 professional sonographers following FASP protocols between
2015 and 2020. The sonographer identified, labelled, and saved standard
plane images and manually measured biometrics. To best reflect the
screening population, all scans were included regardless of whether
the scan was reported as normal or abnormal. 

Although the same machines were used throughout this study, software
updates during this period changed the video resolution and the interface
presented by the machine. Recordings were split across three resolutions
depending on the date on which they were acquired: $678\times576$,
$980\times784$, and $1280\times1024$.

Due to operator error, technical glitches, and patient withdrawal
of consent, not all of these scans could be used in the dataset: 7309
video recordings of prenatal US scans were used for this study.

To maintain patient anonymity, personal information was removed from
the scan recordings and each scan was labelled with a numerical iFIND
ID, numbered sequentially. We created training/validation/test splits
across the dataset assigned subjects using the trailing digit of their
iFIND ID. Scans with an iFIND ID with a trailing digit between 0-5
were used to train our models, 6-7 were used for validation, and 8-9
were held aside for testing and not used during training of any of
our models. The demographic characteristics of patients across these
folds, none of which showed statistically significant differences,
are shown in Table \ref{tab:Patient-metadata}. 

\subsubsection{Validation dataset \label{subsec:iFIND2}}

Some of our methods were also validated using a small number of paired
scans from a smaller study \citep{Matthew2022}. 23 volunteers were
recruited after their 20-week anomaly scan for a follow-up scan. These
ranged in gestational age from 20\textsuperscript{+0} to 24\textsuperscript{+0}
gestational weeks (median 23GW), and had all screened as normal at
their anomaly scan. These subjects had paired scans (encompassing
the FASP planes and biometrics) conducted by separate sonographers
on the same day, using a Philips EpiQ 7 US machine. These scans were
not used to train our models.

\subsection{Data labelling}

\begin{table}[h]

\begin{centering}
\begin{tabular}{cc}
\toprule 
Standard plane & Description\tabularnewline
\midrule
\midrule 
\multicolumn{2}{c}{\textbf{Head:}}\tabularnewline
\midrule 
Brain-CB & Transcerebellar view\tabularnewline
\midrule 
Brain-TV & Transventricular view\tabularnewline
\midrule 
\multicolumn{2}{c}{\textbf{Heart:}}\tabularnewline
\midrule 
3VT & 3-vessel + trachea\tabularnewline
\midrule 
LVOT & left ventricular outflow tract\tabularnewline
\midrule 
RVOT & left ventricular outflow tract\tabularnewline
\midrule 
4CH & 4-chamber view\tabularnewline
\midrule 
\multicolumn{2}{c}{\textbf{Other:}}\tabularnewline
\midrule 
Abdominal & \tabularnewline
\midrule 
Femur & \tabularnewline
\midrule 
Lips & \tabularnewline
\midrule 
Kidneys & axial view\tabularnewline
\midrule 
Profile & sagittal view\tabularnewline
\midrule 
Spine-cor & coronal view\tabularnewline
\midrule 
Spine-sag & sagittal view\tabularnewline
\midrule 
\multicolumn{2}{c}{\textbf{Background}}\tabularnewline
\bottomrule
\end{tabular}
\par\end{centering}
\centering{}\medskip{}
\caption{List of standard planes recommended under FASP. We added a \textquoteleft Background\textquoteright{}
label for frames in none of these views.\label{tab:Standard-planes}}
\end{table}

The iFIND dataset includes full recordings of examinations that were
labelled during scanning with standard planes and biometrics according
to the FASP standard as shown in Table \ref{tab:Standard-planes}
(note there is an additional background class for all other planes).
These labels are present in the recording and were extracted automatically.
Typically, a sonographer freezes the video stream once she identifies
a suitable standard plane. Text labels and biometric measurements
are then overlaid on the frame, which is saved for the sonographer's
report. Example frames with these overlays are shown in Figure \ref{fig:annotation-frames}. 

\subsubsection{Extraction of standard planes}

\begin{figure}[h]
\begin{centering}
\includegraphics[width=0.49\columnwidth]{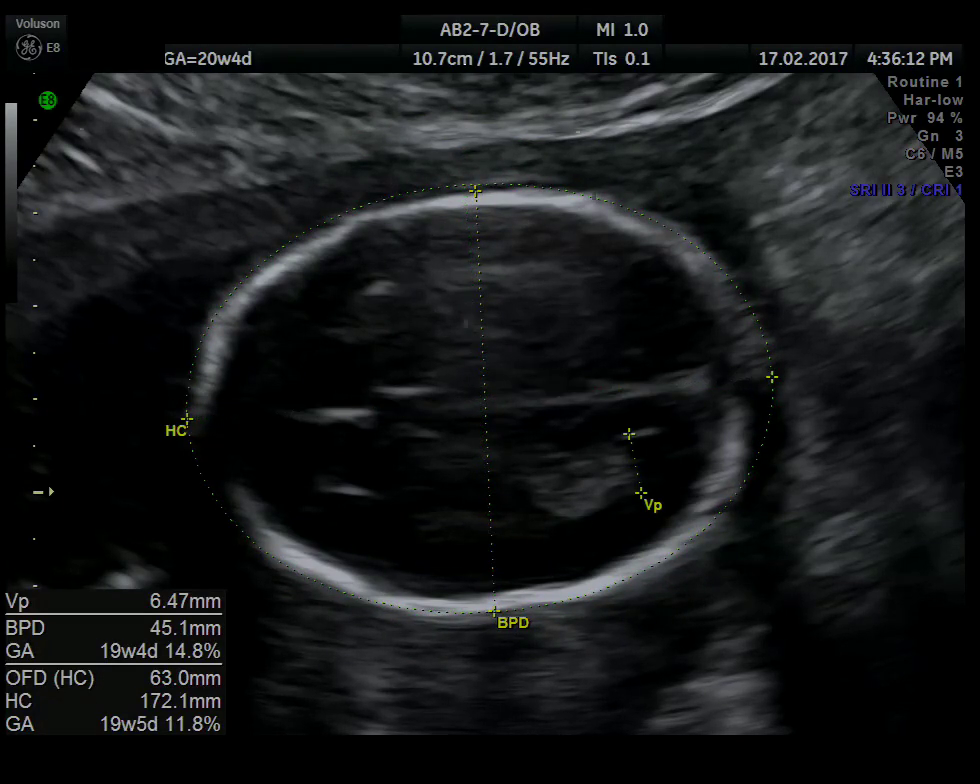}~\includegraphics[width=0.49\columnwidth]{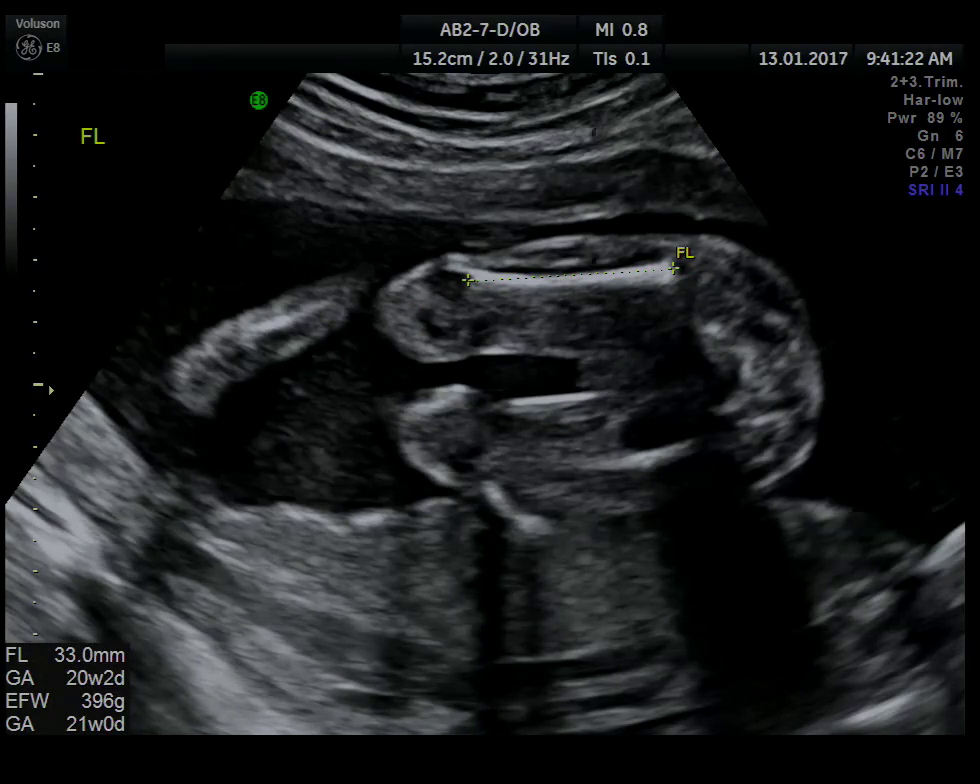}
\par\end{centering}
\begin{centering}
(a)\hspace*{0.45\columnwidth}(b)
\par\end{centering}
\caption{Example frames with annotations and calipers. Shown are (a) a Brain-TV
image, with annotations for head circumference, biparietal diameter
and posterior ventricle, (b) a femur image, with the femur length
measured and annotated. These frames were acquired and saved by a
sonographer. \label{fig:annotation-frames}}
\end{figure}

To find labelled planes in the recorded examinations, we automatically
detected any pauses and freezes in the recordings, then used OCR software\footnote{For this project, we used the open-source Tesseract OCR package \citep{Smith2007}
to read text labels.} to extract any added text labels. The text labels, including biometric
labels (such as the Brain-TV view in Figure \ref{fig:annotation-frames}a)
was then associated with standard planes and manually checked for
consistency. 

In the experiments performed in this paper, we removed segments where
the sonographer froze the frame and added annotations. We did this
by finding frames where over 95\% of pixels did not change relative
to the previous frame. This eliminates the sonographer's own annotations
from the recordings we used to test our methods.

\subsubsection{Biometrics\label{subsec:Biometrics}}

\begin{table*}
\begin{centering}
\begin{tabular}{cccc}
\toprule 
\multicolumn{2}{c}{Biometric} & Type & No. of labels\tabularnewline
\midrule
\midrule 
\multicolumn{3}{l}{\textbf{Brain-TV:}} & \tabularnewline
\midrule 
HC & Head circumference & Ellipse & 8162\tabularnewline
\midrule 
BPD & Biparietal diameter & Ellipse axis & 8162\tabularnewline
\midrule 
\multicolumn{3}{l}{\textbf{Brain-CB:}} & \tabularnewline
\midrule 
TCD & Transcerebellar diameter & Linear & 6906\tabularnewline
\midrule 
\multicolumn{3}{l}{\textbf{Abdominal:}} & \tabularnewline
\midrule 
AC & Abdominal circumference & Ellipse & 6073\tabularnewline
\midrule 
\multicolumn{3}{l}{\textbf{Femur:}} & \tabularnewline
\midrule 
FL & Femur length & Linear & 5718\tabularnewline
\bottomrule
\end{tabular}
\par\end{centering}
\begin{centering}
\medskip{}
\caption{Biometrics measured during the 20-week scan according to FASP guidelines,
and the number of such labels present in the iFIND1 dataset. \label{tab:Biometrics}}
\par\end{centering}
\end{table*}

In the UK, the FASP standard mandates a minimum of 5 biometric measurements
per scan \citep{FASP}\footnote{Other biometrics, such as the nuchal fold (NF), must be measured in
certain circumstances but are not required for every scan.}. These are measured across a range of standard planes, as shown in
Table \ref{tab:Biometrics}. 

\begin{figure}[h]
\begin{centering}
\includegraphics[width=0.48\columnwidth]{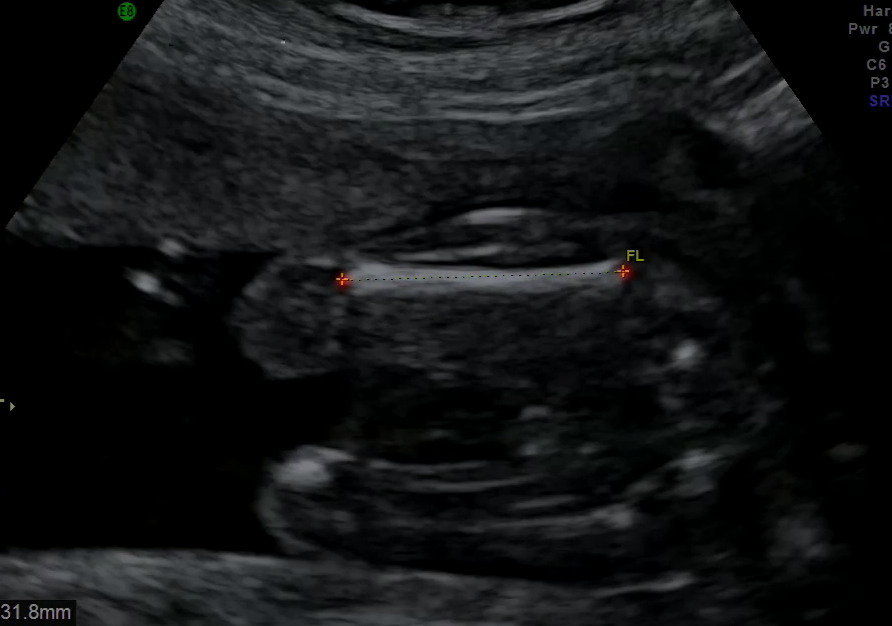}~\includegraphics[width=0.48\columnwidth]{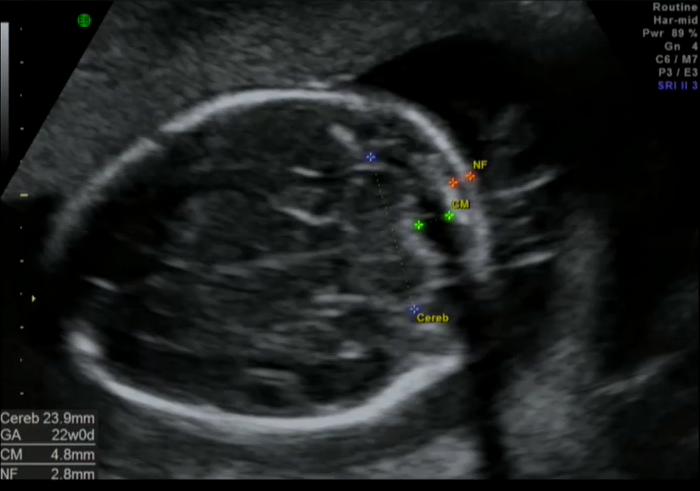}
\par\end{centering}
\begin{centering}
(a)\hspace*{0.45\columnwidth}(b)
\par\end{centering}
\caption{CaliperNet example heatmap outputs for (a) a femur image, and (b)
a brain-CB image with labels for the cerebellum, cisterna magna and
nuchal fold. \label{fig:CaliperNet}}
\end{figure}

The calipers identifying a biometric are generally connected by a
dotted line, with a text label next to them to identify the calipers
(see Figure \ref{fig:CaliperNet}). To extract these biometrics, we
trained a simple U-Net based neural network which we called CaliperNet
(see Appendix \ref{sec:CaliperNet} for details). We manually labelled
300 images of each biometric with the location of the calipers and
trained a CNN to find calipers, with very good accuracy.

\subsection{Pixel size estimation}

\begin{figure*}
\begin{centering}
\includegraphics[viewport=14.40625bp 25.19043bp 36.0156bp 165.5371bp,clip,width=0.053\textwidth]{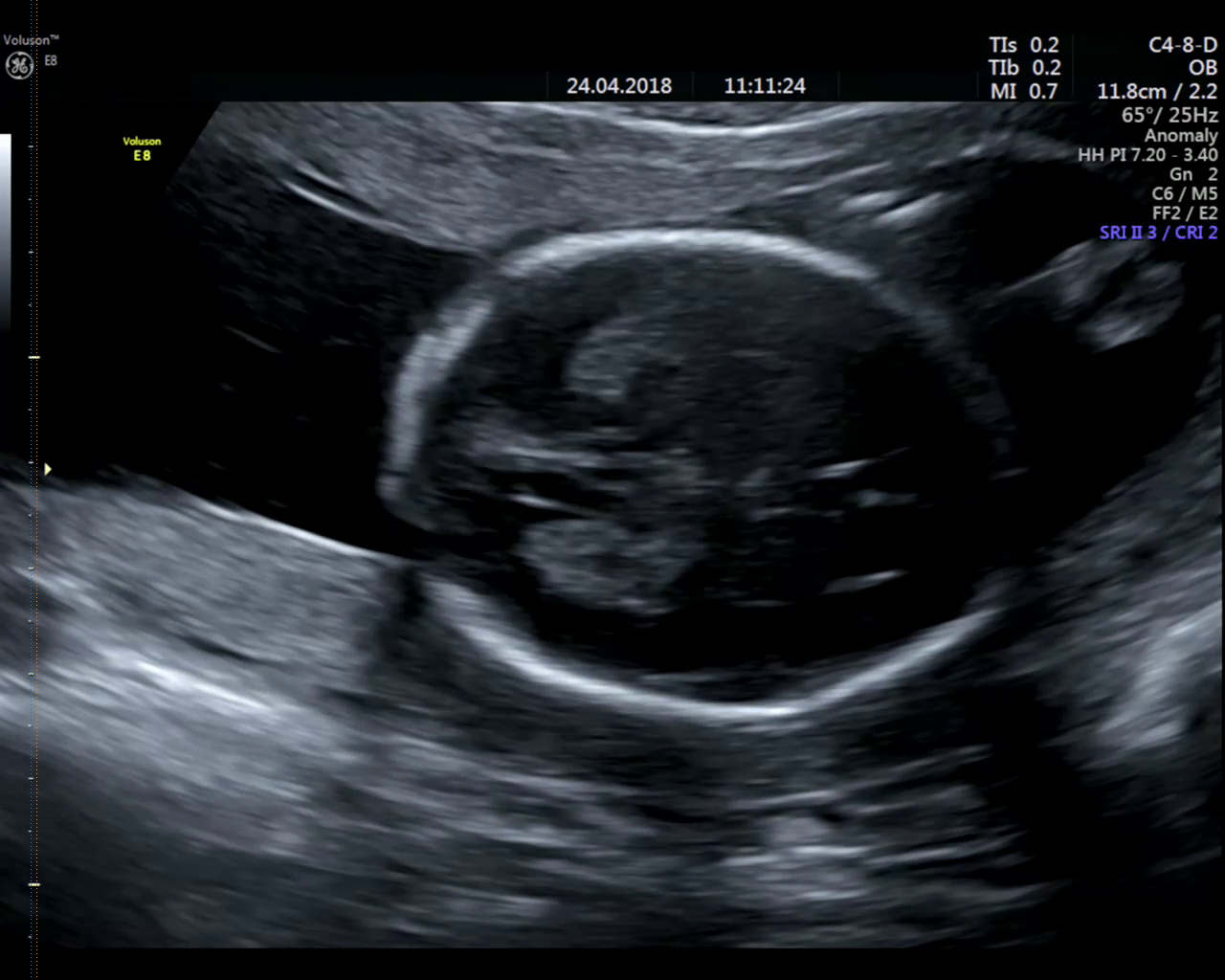}\includegraphics[width=0.47\textwidth]{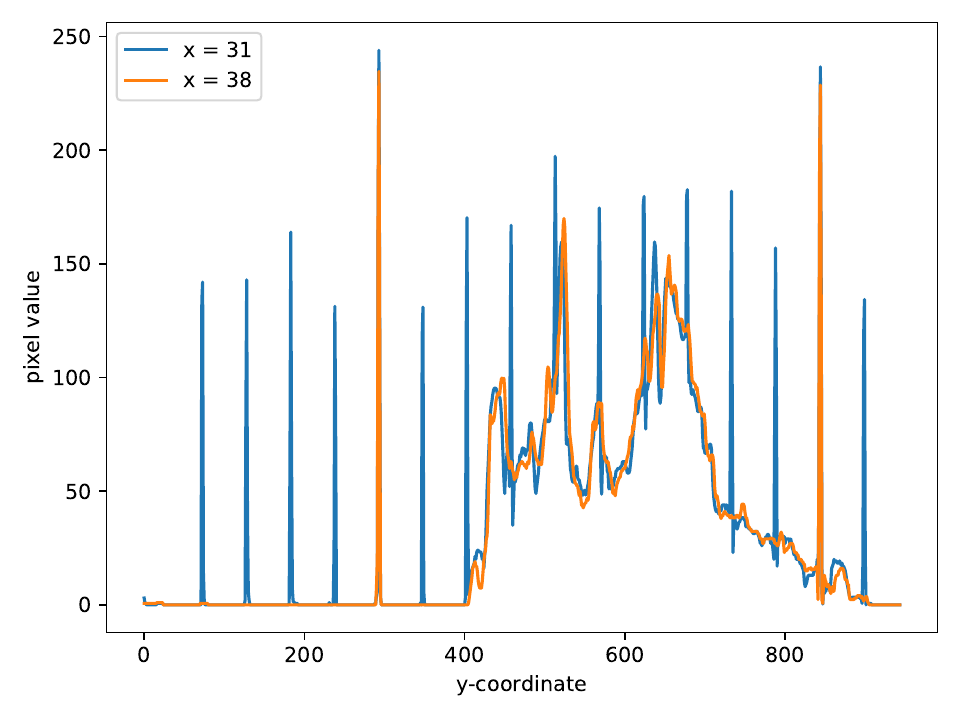}\includegraphics[width=0.47\textwidth]{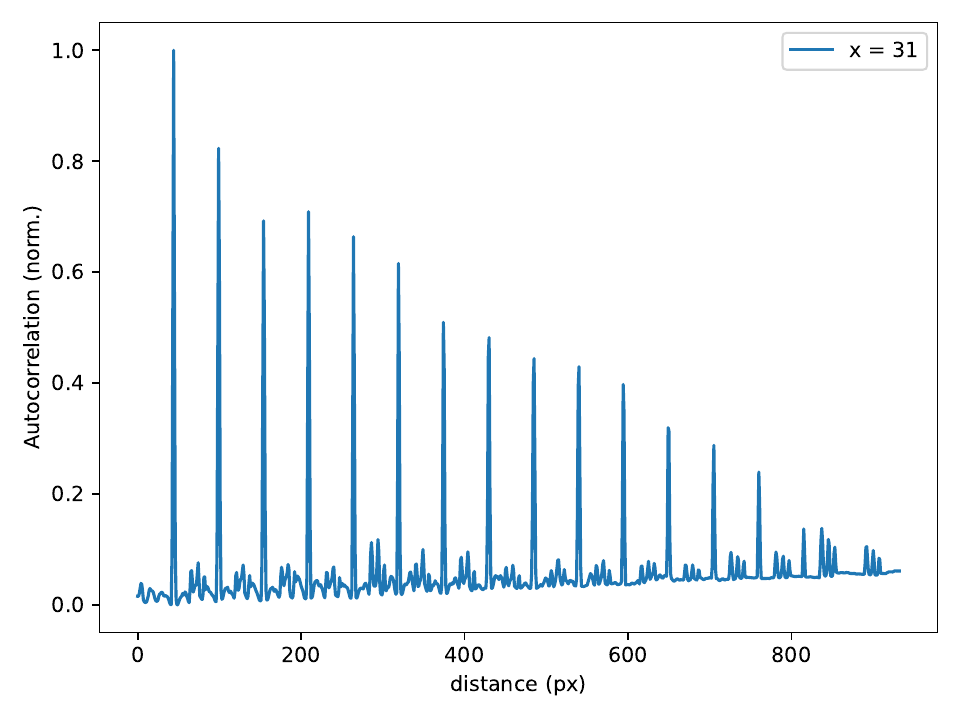}
\par\end{centering}
\begin{raggedright}
\hspace*{0.02\textwidth}(a)\hspace*{0.25\textwidth}(b)\hspace*{0.45\textwidth}(c)
\par\end{raggedright}
\caption{(a) Part of a scale bar, cropped from an image frame. Two lines are
overlaid upon it. (b) The pixel values along those two scan lines
for the entire frame. (c) The autocorrelation of the pixel values
of the blue line after it has been processed with a high-pass filter.\label{fig:Pixel-size-estimation}}
\end{figure*}

The method described in Section \ref{tab:Biometrics} can find biometric
endpoints on an image, but the measurements are given in pixels rather
than millimetres. Moreover, pixel size can be changed during scanning
by varying probe settings and machine zoom levels.

However, almost every frame shows a standard scale bar with prominent
ticks at intervals $d_{\mathrm{bar}}=50$mm, with smaller ticks every
10mm and 5mm \citep{VolusonManual2012}. These ticks are at predictable
positions on the screen: they can be seen clearly by looking at pixel
values $X_{i}$ along the correct scan line (see Figure \ref{fig:Pixel-size-estimation}).
Ticks can sometimes blend into bright structures shown on screen. 

A simple, reliable and computationally efficient solution is to remove
likely background signals from $X_{i}$ by subtracting a 1D Gaussian
filter, $G_{x},$with standard deviation $\sigma$ empirically selected
to be 3px and keeping only positive values:

\begin{equation}
X_{i}^{'}=\max\left(X_{i}-X_{i}\ast G_{x}(\sigma),0\right),
\end{equation}
We then computed the autocorrelation $R_{XX}$ of the resulting filtered
pixel values $X_{i}^{'}$ along the scan lines corresponding to the
scale bar (see Figure \ref{fig:Pixel-size-estimation}): 
\begin{equation}
R_{XX}(n)=\sum_{i}X_{i}^{'}X_{i+n}^{'}.
\end{equation}
The first peak of the autocorrelation function corresponds to the
spacing between axis ticks (in pixels). The size of individual pixels
in millimetres $L_{x}$ is given by
\begin{equation}
L_{x}=\frac{d_{\mathrm{bar}}}{\max(R_{XX}(n))}.
\end{equation}

To minimise quantisation noise, we estimate pixel size from the bars
with the highest spacing.

\subsection{Biometric measurement \label{subsec:Biometric-measurement}}

\begin{figure}[h]
\begin{centering}
\textbf{\includegraphics[width=0.48\columnwidth]{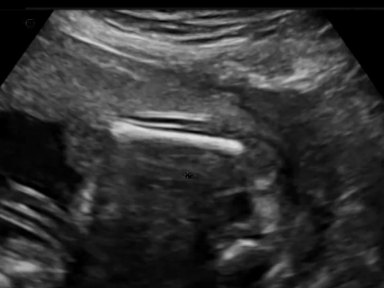}~\includegraphics[width=0.48\columnwidth]{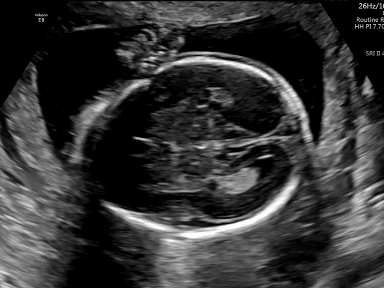}\medskip{}
\includegraphics[width=0.48\columnwidth]{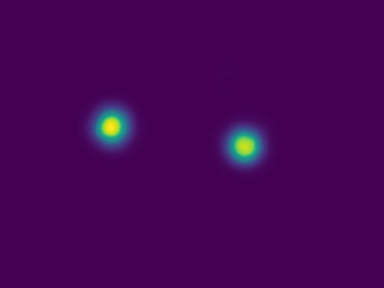}~\includegraphics[width=0.48\columnwidth]{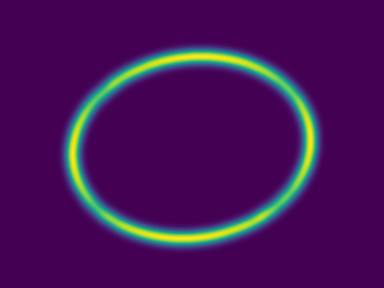}\medskip{}
\includegraphics[width=0.48\columnwidth]{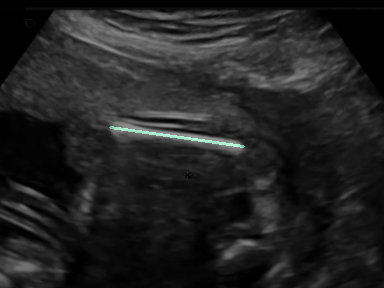}~\includegraphics[width=0.48\columnwidth]{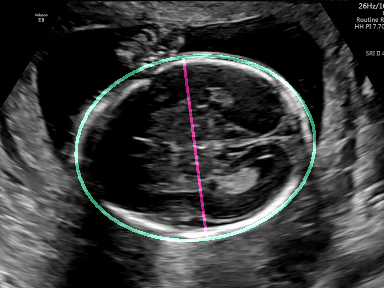}}
\par\end{centering}
\begin{centering}
(a)\hspace*{0.45\columnwidth}(b)
\par\end{centering}
\caption{Images used for biometric training showing (a) femur, and (b) head
views. The second row shows the training labels and the third row
shows the CNN output overlaid on the image. For the head image, the
HC output is shown in green and the BPD output in pink.\label{fig:Paired-image}}

\end{figure}

To measure biometrics, we trained a the U-Net segmentation network
using binary cross-entropy loss \citep{Sinclair2018} to predict heatmaps
constructed from the sonographer\textquoteright s point annotations,
extracted by CaliperNet, convolved with a Gaussian kernel. Figure
\ref{fig:Paired-image} shows examples of training labels constructed
this way. A separate CNN was trained for each standard plane, using
the same U-Net architecture for each network and differing only in
training planes and labels.

Biometrics were extracted from unseen images directly from the predicted
heatmaps. For linear biometrics (FL and TCD), the Euclidean distance
between the coordinates of the two highest maxima was used; For elliptical
biometrics (HC and AC) we fit an ellipse to the heatmap using the
least-squares criterion. The ellipse perimeter was then estimated
using the formula recommended by the British Medical Ultrasound Society\footnote{The exact calculation of the perimeter of an ellipse is an infinite
series. The cited guideline suggests using the first five terms of
this series, which has a relative error of $<0.01\%$ for most biologically
plausible ellipse shapes.} \citep{BMUS}.

Since BPD is mathematically equivalent to the minor axis of the ellipse
used to segment the head, we calculated it directly from the head
circumference label.

\subsection{Whole-scan biometric estimation \label{subsec:Whole-video-biometric-estimation}}

In clinical practice, only a few frames are labelled by the sonographer.
Generally, no more than three measurements are taken of each biometric
\citep{FASP} and the best subjective measurement is reported.

\begin{figure}
\begin{centering}
\includegraphics[width=1\columnwidth]{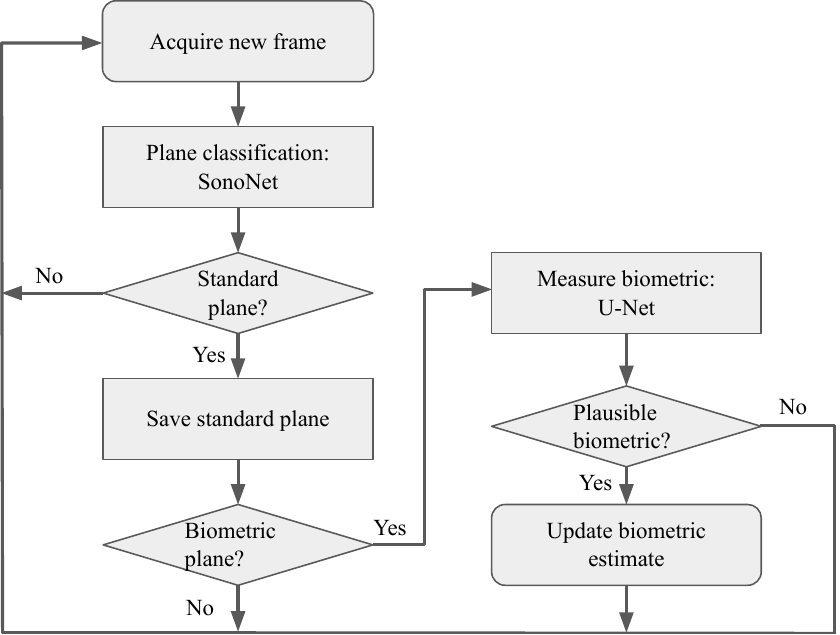}
\par\end{centering}
\caption{The pipeline which each frame in an US scan feed is processed by to
estimate biometrics.\label{fig:Pipeline}}
\end{figure}

The automated methods described in Section \ref{subsec:Biometric-measurement}
can extract measurements in all identified frames in which a given
biometric is visible, resulting in hundreds or thousands of measurements
per biometric. These form sample populations from which we wish to
infer a true value and to estimate confidence intervals. 

However, for each biometric there may be erroneous values, caused
by misclassification of the target anatomical planes or by failure
modes of the biometric CNN. Any estimation needs to reject or at least
minimise the effect of these out-of-distribution measurements to reliably
converge on the true biometric value.

Figure \ref{fig:Pipeline} shows our processing pipeline. Every frame
from the US real time image stream is processed by Sononet to identify
and label standard planes. Following Baumgartner et al \citep{Baumgartner2017}:,
each frame is resized to a resolution of $288\times224$ and converted
to greyscale.

Any frame identified with $>95\%$ confidence as a standard plane
with a FASP biometric is then processed by our CNNs to obtain a biometric
measurement. The native frame is subsampled to a $384\times288$ resolution,
with interface elements removed, and converted to greyscale and processed
as described in Section \ref{subsec:Biometric-measurement}. 

A quick check is then applied: if the proposed measurement is anatomically
implausible (lower than the 3\textsuperscript{rd} centile of the
biometric at $\mathrm{GA}-3$ weeks or higher than the 97\textsuperscript{th}
centile at $\mathrm{GA}+3$ weeks), it is rejected. Furthermore, where
the CNN's output does not allow an ellipse to be fitted or endpoints
to be returned (for instance, where one distinct maximum is found,
making it impossible to find two endpoints) the output is discarded. 

These two filters (of plane-classification confidence and biometric
measurements) may be undesirable for single-frame comparisons, as
they discard some data. However, when estimating biometrics across
a whole scan several hundred measurements of each biometric are often
recorded: if some fraction of these are discarded, this only has a
small impact on data availability. Appendix \ref{sec:techreq-dump}
details the proportion of biometric measurements discarded. 

\subsubsection{Distribution of measurements}

\begin{figure}
\centering{}\includegraphics[width=1\columnwidth]{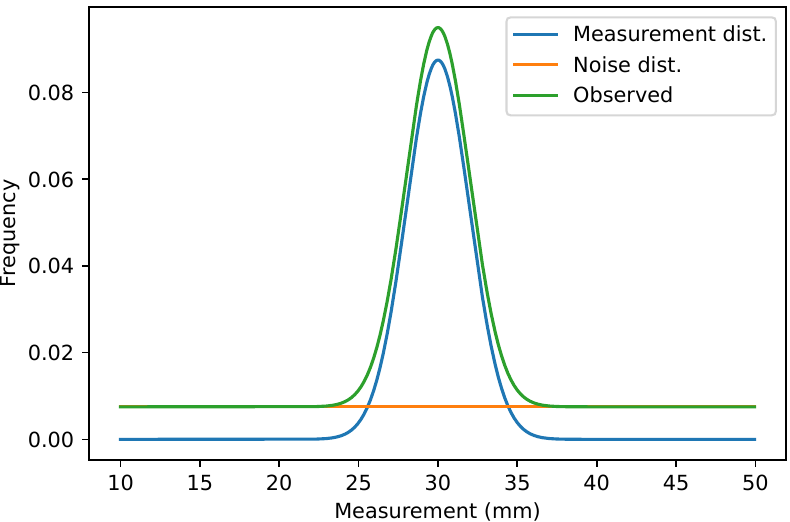}\caption{Idealised distribution of measurements for a biometric with true value
30mm. The observed distribution of measurements is a sum of a measurement
distribution and a nuisance distribution of outliers.\label{fig:Distribution-graph}}
\end{figure}

This section describes how measurements of individual frames are aggregated
and outliers are rejected using a Bayesian framework.

Repeated measurements of the same biometric in different frames should
cluster around a mean value. We modelled this using a normal distribution
$\mathcal{N}\left(\mu,\sigma^{2}\right)$, with mean $\mu$ (if unbiased,
this should be the true biometric value) and variance $\sigma^{2}$(this
depends on the precision of the biometric measurements).

Misclassified planes also return a biometric measurement and the biometric
measurement process itself may fail even in a correctly identified
plane. Since we have no information with which to model these failures
we treat them as random and use a uniform distribution $U\left(a,b\right)$,
over the full biological limits $a$ and $b,$ outside of which a
measurement is rejected.

Thus, the observed distribution $D_{o}$ of measurements will be a
weighted sum of the measurement distribution $\mathcal{N}\left(\mu,\sigma^{2}\right)$
(for correct classifications) and a nuisance distribution $U\left(a,b\right)$
(for misclassifications). This can be represented by
\begin{equation}
D_{o}=P_{t}\mathcal{N}\left(\mu,\sigma^{2}\right)+(1-P_{t})U\left(a,b\right),
\end{equation}
where $P_{t}$ represents the proportion of valid measurements.

This equation contains several unknowns $P_{t},\mu,\sigma^{2}$ which
can be estimated iteratively using a Bayesian approach, but each must
be initialised.

\subsubsection{Estimating truth probability and distribution parameters}

To perform a Bayesian estimation, $P_{t}$needs to be initialised
to a prior probability (in our approach, we set $P_{t_{0}}=0.75$
for all biometrics, but it should converge on the true probability
from any starting point other than 0 or 1), along with a prior weight
$W_{0}$\footnote{The estimate should converge on the true value of $P_{t}$ regardless
of what $W_{0}$ is set to, but setting it too high can cause it to
converge slowly, while setting it too low can cause the estimate of
$P_{t}$ to oscillate inappropriately.}. Prior estimates of the distribution parameters $\mu_{0}$ and $\sigma_{0}^{2}$
need to be set, along with appropriate prior weights $W_{\mu,0}$
and $W_{\sigma^{2},0}$.

For each new measurement $x_{i}$, we estimate the likelihood of it
being a true measurement $T$ or a nuisance sample $\overline{T}$
using Bayes' rule: 
\begin{equation}
P(T|x_{i})=\frac{P(x_{i}|T)P_{t}}{P(x_{i}|\overline{T})(1-P_{t})}
\end{equation}

where $P(x_{i}|T)$ is the value of $\mathcal{N}\left(\mu,\sigma^{2}\right)$
at $x_{i}$, and $P(x_{i}|\overline{T})$ is the value of $U\left(a,b\right)$
at value $x_{i}$. This returns $P(T|x_{i}),$ the probability that
this measurement was sampled from the true measurement distribution.

Finally, $P_{t}$ and the distribution parameters $\mu,\sigma^{2}$
can be updated for this biometric using a weighted cumulative average:

\begin{eqnarray}
P_{t_{i}} & = & \frac{P(T|x_{i})+W_{i-1}P_{t_{i-1}}}{W_{i-1}+1}\\
\mu_{i} & = & \frac{x_{i}P(T|x_{i})+W_{\mu,i-1}\mu_{i-1}}{P(T|x_{i})+W_{\mu,i-1}}\\
\sigma_{i}^{2} & = & \frac{(x_{i}-\mu_{i-1})^{2}P(T|x_{i})+W_{\sigma^{2},i-1}\sigma_{i-1}^{2}}{P(T|x_{i})+W_{\sigma^{2},i-1}}
\end{eqnarray}

and their weighting factors can be updated using

\begin{eqnarray}
W_{i} & = & W_{i-1}+1\\
W_{\mu,i} & = & W_{\mu,i-1}+P(T|x_{i})\\
W_{\sigma^{2},i} & = & W_{\sigma^{2},i-1}+P(T|x_{i}).
\end{eqnarray}

As the noise distribution is taken to be uniform, there is no need
to update estimates of its terms.

The estimates for $\mu$ and $\sigma^{2}$ are updated in real time
and independently. At any given moment, the estimate of $\mu$ is
the best estimate of the true value of the biometric of interest.
Meanwhile, the estimate of $\sigma^{2}$ can be used to calculate
the standard error ,$\hat{\sigma}_{i}$, of that estimate to find
credible intervals for the biometric, given by
\begin{equation}
\hat{\sigma}_{i}=\frac{\sigma_{i}}{\sqrt{W_{\sigma^{2},i}}}.
\end{equation}
The 95\% credible interval is given by the range $\mu_{i}\pm2\hat{\sigma}_{i}$.

\section{Experiments}

The variability in biometric measurements between two sonographers
arises from two components: image selection for measurement and caliper
placement within each image. Previous work has shown that 50-80\%
of variability between humans can be explained by caliper placement
\citep{Sarris2012}, leaving 20-50\% of variability accounted for
by image selection. The proposed whole-scan method eliminates this
second source of noise by processing every frame in a scan with no
manual intervention and seeks to reduce the effect of random error
in caliper placement by aggregating measurements from many individual
frames.

We performed three experiments to quantify the reliability of our
biometric estimates:
\begin{enumerate}
\item Using our biometric CNNs to independently measure the same frames
that sonographers labelled. This tests the biometric measurement performance
of our CNNs and compares it to published estimates of inter-rater
differences in caliper placement on a single image. Note this is still
a restricted problem, requiring human intervention to perform frame
selection. For these single-image experiments, biometrics were estimated
automatically in the same frames that sonographers chose to make their
manual annotations. We trained one CNN per standard plane (from subjects
in the training set) and tested it using the subjects in the test
set using the unannotated copies previously extracted for all frames
in the iFIND test set: 1516 in the brain-TV standard plane, 1360 for
brain-CB, 1205 for abdominal, and 1124 for femur.
\item Aggregating biometric measurements across a whole recorded scan to
obtain global estimates for the scan and comparing to sonographer
manual measurements. This experiment considers a more unconstrained
problem with no human interaction. It does not control for individual
frame selection, so our results can be compared to inter-rater variability
in biometric measurement from different scans. All 1457 recordings
in the test set were used. The final estimate of each biometric, along
with credible intervals, was reported. 
\item Conducting a test-retest experiment on the paired scan data described
in Section \ref{subsec:iFIND2}. Finally, we conducted a test-retest
experiment on the paired scan data described in Section \ref{subsec:iFIND2}.
We ran the full pipeline on paired scans of the same subject acquired
on the same day using a different scanner from the training dataset.
We then compared biometrics from each scan and measured test-retest
variability of our algorithm. The scans here were performed on an
US machine by a different manufacturer from the training data, which
allowed us to examine domain shift.
\end{enumerate}
All experiments were conducted on a single Nvidia GeForce RTX 3080
GPU. 

\subsection{Validation of distribution choice}

Section \ref{subsec:Whole-video-biometric-estimation} models the
distribution of measurements of each biometric with a mixture of a
uniform distribution and a Gaussian distribution 
\begin{equation}
D=P_{t}\mathcal{N}\left(\mu,\sigma^{2}\right)+(1-P_{t})U\left(a,b\right).
\end{equation}

This has three parameters $P_{t},\mu$ and $\sigma$ which are tuned
iteratively to incoming data. This model assumes that incoming measurements
follow a Gaussian distribution (if correctly classified) or a uniform
distribution (if noise), which can be tested. We therefore examined
the output biometric measurements to check the appropriateness of
this model.

\section{Results}

\subsection{Single-frame biometric estimation\label{subsec:Single-frame-biometric-estimatio}}

\begin{table*}
\begin{centering}
\begin{tabular}{cccc}
\toprule 
Structure & Bias (mm) & MSD (mm) & Human MSD\tabularnewline
\midrule
\midrule 
HC & +0.11 mm (+0.07\%) & 4.34mm (2.68\%) & 2.70mm (1.8\%)\tabularnewline
\midrule 
BPD & +0.06mm (+0.17\%) & 1.28mm (3.20\%) & N/A\tabularnewline
\midrule 
AC & +1.51mm (+0.98\%) & 5.03mm (3.26\%) & 4.03mm (2.8\%)\tabularnewline
\midrule 
FL & 0.00mm (0.00\%) & 1.53mm (4.67\%) & 1.02mm (2.9\%)\tabularnewline
\midrule 
TCD & -0.16mm (-0.75\%) & 2.1mm (10.1\%) & N/A\tabularnewline
\bottomrule
\end{tabular}\medskip{}
\par\end{centering}
\caption{Comparison of the biometric measurements obtained on the test set
of the iFIND dataset. \textquoteleft Bias\textquoteright{} refers to
the average difference in measurement between our method and human
sonographers. \textquoteleft MSD\textquoteright{} is the mean-squared
difference between pairs of measurements. The \textquoteleft Human
MSD\textquoteright{} column is populated with values of inter-observer
variability reported by Sarris et al \citep{Sarris2012}. \label{tab:Frame-by-frame-results}}
\end{table*}

\begin{figure}
\begin{centering}
\includegraphics[width=0.49\columnwidth]{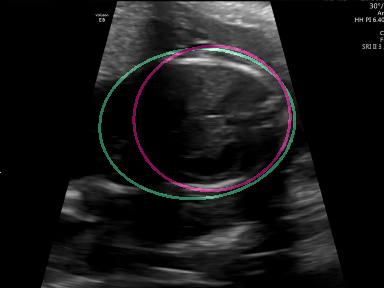}~\includegraphics[width=0.49\columnwidth]{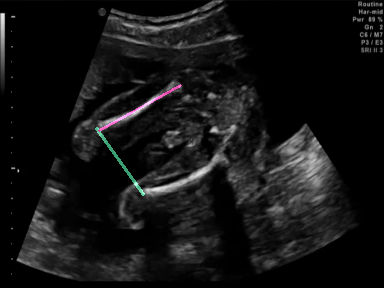}
\par\end{centering}
\begin{centering}
(a)\hspace*{0.45\columnwidth}(b)
\par\end{centering}
\caption{Two failure cases of biometric estimation in (a) HC, and (b) FL. The
pink annotation shows the ground truth measurement made by a sonographer,
while the green annotation is the automatic one made by our CNNs.\label{fig:Failure-frames}}
\end{figure}

\begin{table*}[t]
\begin{centering}
\begin{tabular}{cccccc}
\toprule 
Biometric & Bias & MSD & MAD (mm) & std. err (mm) & within human 95\%\tabularnewline
\midrule
\midrule 
HC & -0.30mm (-0.18\%) & 3.23mm (1.99\%) & 2.50mm (1.42\%) & 3.39mm & 94.7\%\tabularnewline
\midrule 
BPD & +0.90mm (+1.89\%) & 1.86mm (4.65\%) & 1.11mm (2.31\%) & 0.95mm & N/A\tabularnewline
\midrule 
AC & -0.49mm (-0.31\%) & 5.55mm (3.60\%) & 4.23mm (2.74\%) & 3.26mm & 96.6\%\tabularnewline
\midrule 
FL & -0.51mm (-1.59\%) & 1.63mm (4.99\%) & 1.19mm (3.64\%) & 1.12mm & 93.9\%\tabularnewline
\midrule 
TCD & +0.27mm (+1.32\%) & 0.91mm (4.37\%) & 0.68mm (3.28\%) & 0.47mm & N/A\tabularnewline
\bottomrule
\end{tabular}
\par\end{centering}
\medskip{}

\caption{Comparison of human measurements to machine estimates taken across
a whole scan. \textquoteleft MAD\textquoteright{} refers to mean absolute
difference between human and machine measurements. The \textquoteleft within
human 95\%\textquoteright{} column shows the percentage of biometrics
in the test set where the human-machine difference is within 95\%
of human-human inter-observer differences (taken from \citet{Sarris2012})\label{tab:Test-retest-performance}}
\end{table*}

A valid biometric estimate (following the constraints outlined in
Section \ref{subsec:Whole-video-biometric-estimation}) was obtained
in approximately 90\% of cases across all biometrics. Table \ref{tab:Frame-by-frame-results}
shows performance on the unlabelled copies of the frames labelled
by sonographers in the test dataset. 

Using the sonographer measures as the reference point, the automated
single frame performance showed bias of less than 1\% and mean square
difference (MSD) of less than 5\% except for TCD which showed a MSD
of just over 10\%. Where inter-observer data is available for manual
measurements, the automated approach could be seen to have an average
of 41\% lower agreement with the human rater.

\subsection{Whole-scan processing\label{subsec:Whole-video-processing}}

\begin{figure}[h]
\begin{centering}
\includegraphics[width=1\columnwidth]{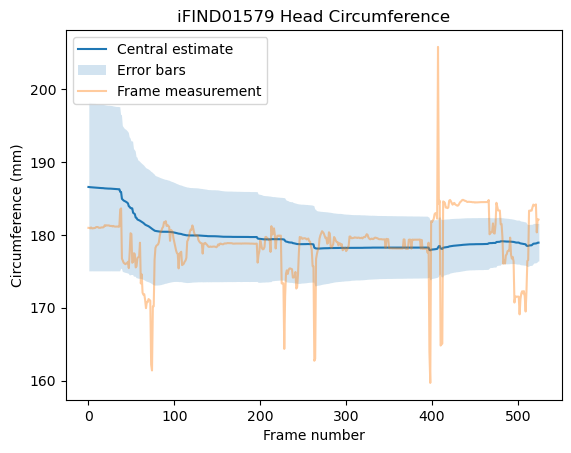}
\par\end{centering}
\medskip{}

\caption{Evolution over time of the HC estimate in an individual subject, overlaid
with credible intervals and individual frame measurements.\label{fig:Biometric-evolution}}
\end{figure}

\begin{figure*}
\begin{centering}
\includegraphics[width=1\textwidth]{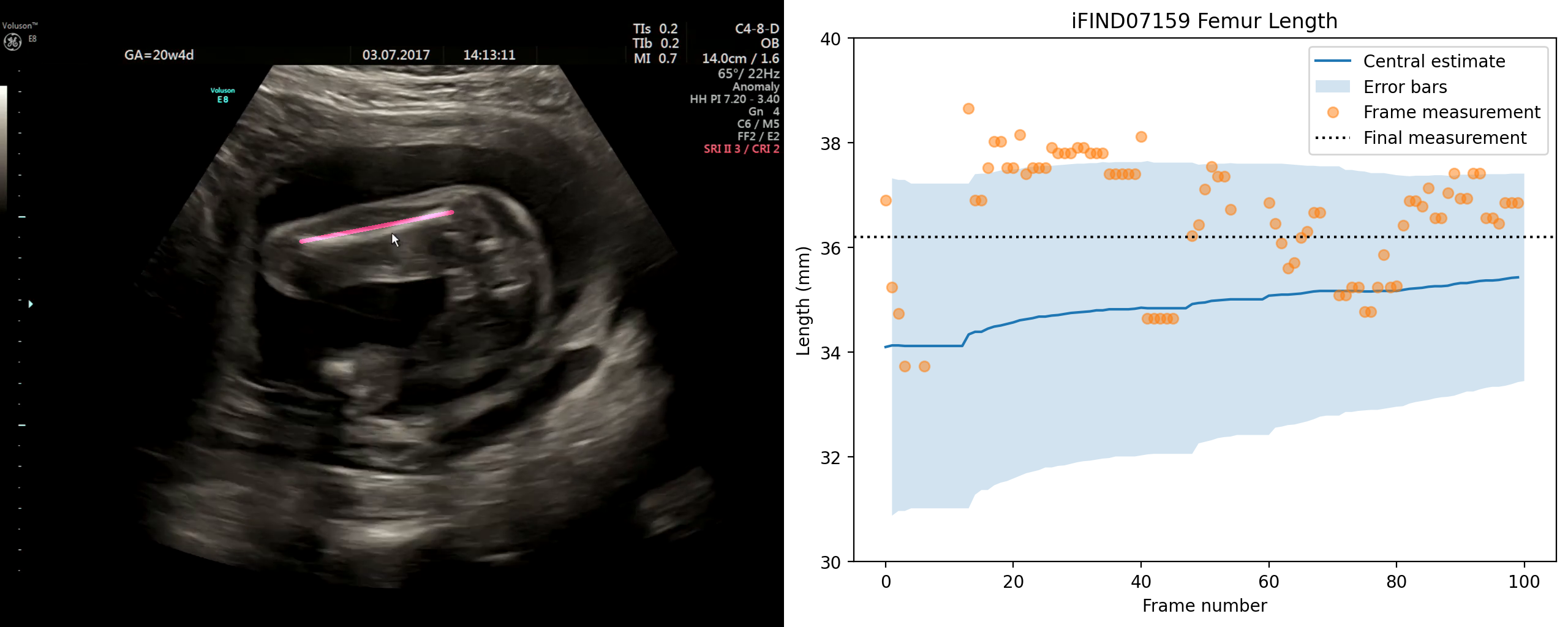}
\par\end{centering}
\caption{A still image from a video showing real-time measurement of biometrics
and updating of the biometric estimate. In three seconds of real-time
scanning, our system acquired 83 FL measurements and updated its FL
estimate from 34mm (the prior, based on the average FL for this gestational
age) to 36mm. The full video clip can be found at \protect\href{http://www.ifindproject.com/wp-content/uploads/sites/79/2023/12/realtime.gif}{http://www.ifindproject.com/wp-content/uploads/sites/79/2023/12/realtime.gif}
\label{fig:video}}
\end{figure*}

\begin{figure*}
\begin{centering}
\includegraphics[width=0.5\textwidth]{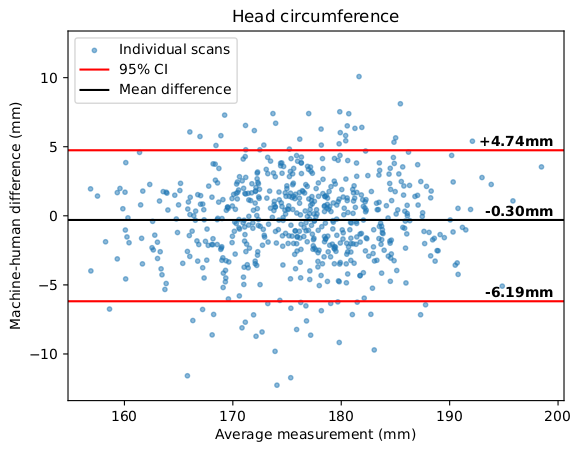}\includegraphics[width=0.5\textwidth]{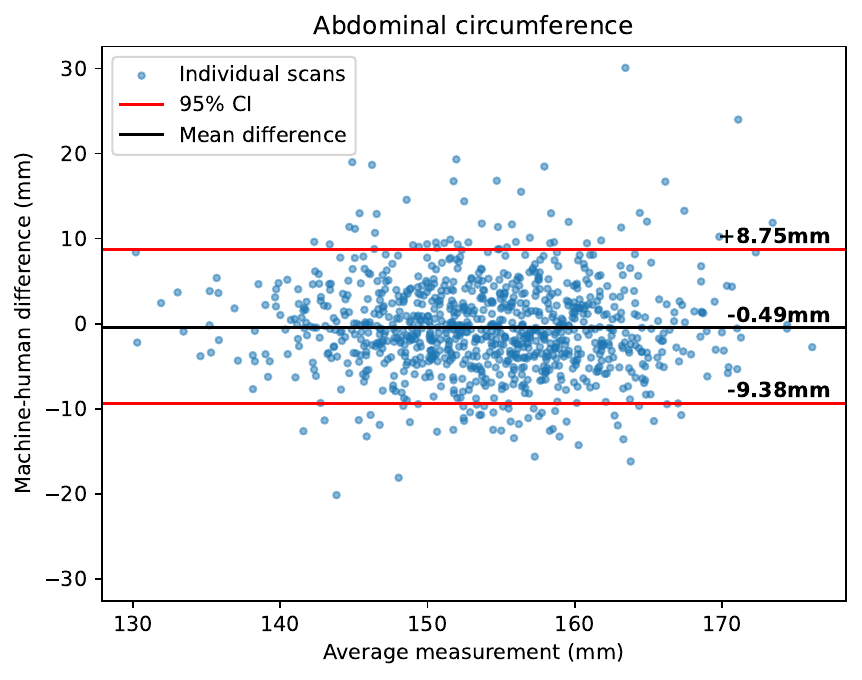}
\par\end{centering}
\begin{centering}
\includegraphics[width=0.5\textwidth]{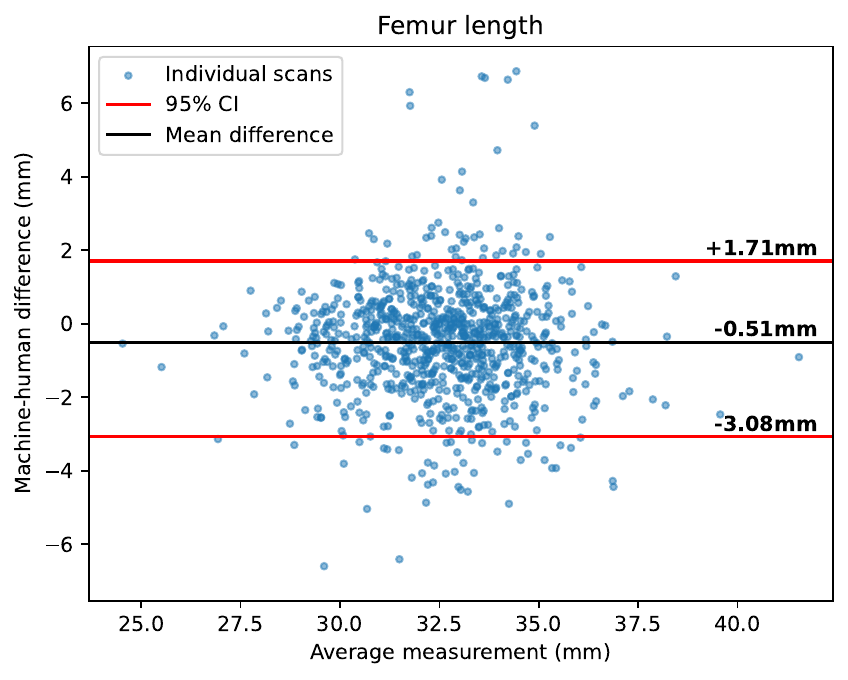}\includegraphics[width=0.5\textwidth]{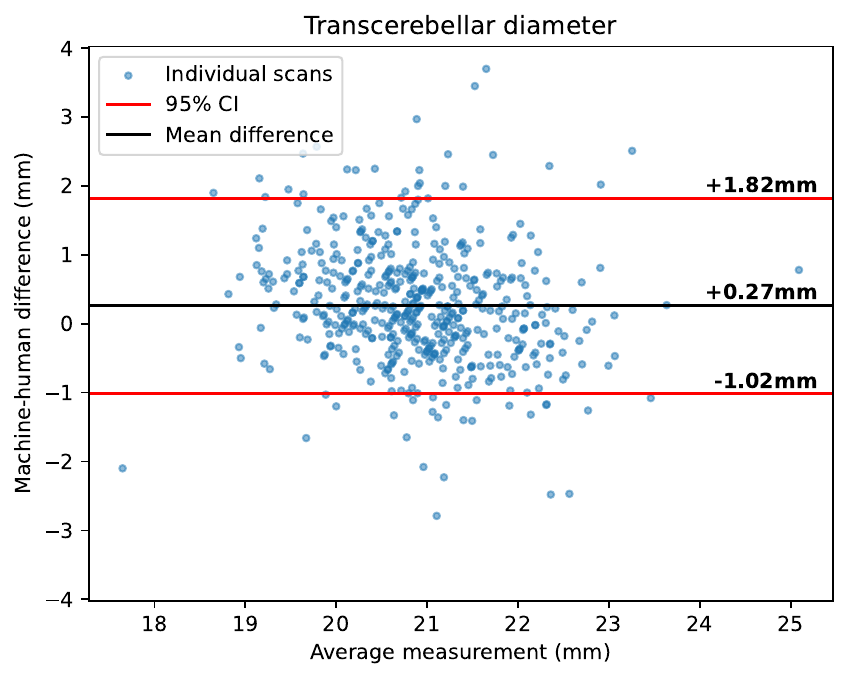}
\par\end{centering}
\caption{Bland-Altman plots showing the differences between the sonographer's
measurements and our own across all complete recordings in the test
dataset.\label{fig:Bland-Altman-plots-for}}
\end{figure*}

Figure \ref{fig:Biometric-evolution} shows how a biometric estimate
(in this case HC) changes over time during one exemplar scan, overlaid
with measurements from individual frames. Initially the prior dominates,
estimating the 50\textsuperscript{th} percentile for the age with
wide error bars. As the number of frames contributing measurement
values increases the overall estimate becomes more stable and resistant
to outliers. Although individual frame measurements show significant
noise, including extreme values (which are probabilistically discounted
by our system), the credible interval progressively shrinks with increasing
data. Figure \ref{fig:video} shows this happening in real time over
3 seconds of scanning.

Figure \ref{fig:Bland-Altman-plots-for} shows Bland-Altman plots
for human and machine measurement on the same subjects for three biometrics,
with 95\textsuperscript{th} percentile intervals overlaid for three
biometrics. The difference between human measurements\footnote{Human measurements were taken from sonographers' reports after each
scan.} and machine estimates can be compared to inter-rater disagreement
measured by Sarris et al \citep{Sarris2012}. Our 95\textsuperscript{th}
percentile intervals are overall very similar to human inter-rater
variability for HC, AC, and FL. 95.1\% of machine-human measurement
differences (across these three biometrics) lie within the 95\% range
of human differences.

\subsection{Paired scan data \label{subsec:Paired-scan-data}}

\begin{table*}
\centering{}%
\begin{tabular}{ccccc}
\toprule 
Biometric & MSD & Standard deviation & Human s.d. (full scan) & Number of pairs\tabularnewline
\midrule
\midrule 
HC & 2.60mm (1.25\%) & 1.23\% & 2.75\% & 15\tabularnewline
\midrule 
BPD & 0.79mm (1.36\%) & 1.17\% & N/A & 15\tabularnewline
\midrule 
AC & 4.61mm (2.51\%) & 2.50\% & 5.05\% & 17\tabularnewline
\midrule 
FL & 0.97mm (2.44\%) & 2.44\% & 5.56\% & 10\tabularnewline
\midrule 
TCD & 0.57mm (2.33\%) & 1.62\% & N/A & 8\tabularnewline
\bottomrule
\end{tabular}\medskip{}
\caption{Test-retest reliability of our model outputs from paired scans of
the same subject. The \textquoteleft Human s.d.\textquoteright{} column
is populated with results by Sarris et al \citep{Sarris2012}. \label{tab:Test-retest-reliability-of}}
\end{table*}

We also conducted a test-retest experiment to measure reliability
across repeated scans of the same subject. 

Table \ref{tab:Test-retest-reliability-of} shows the mean-square
difference (MSD) and the standard deviation of the difference in the
final biometric estimates from the test-retest experiment using data
from 20 subjects scanned twice on the same day by different sonographers,
as described in Section \ref{subsec:iFIND2}\footnote{We measured percent standard deviation in this table to ensure comparability
with the measurements reported by Sarris et al \citep{Sarris2012},
who reported human standard deviation. They examined a slightly different
gestational age range, so we report percent standard deviation to
normalise for that.}. Not all pairs of scans had sufficient frames visible of each biometric
in both scans: the number of pairs available for analysis is shown
in the table.

\subsection{Measurement distributions\label{subsec:Measurement-distributions-results}}

\begin{figure}[h]

\begin{centering}
\includegraphics[width=0.5\columnwidth]{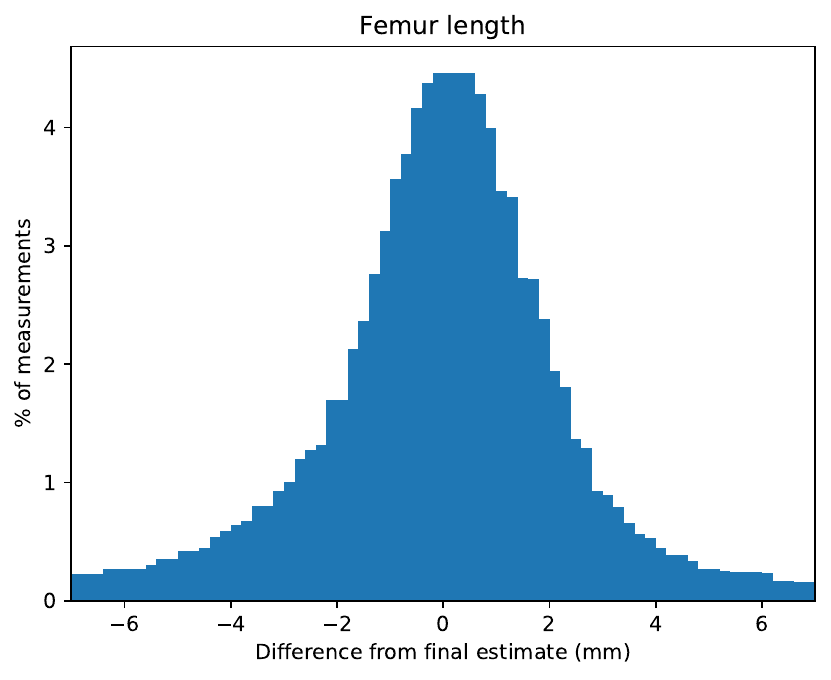}\includegraphics[width=0.5\columnwidth]{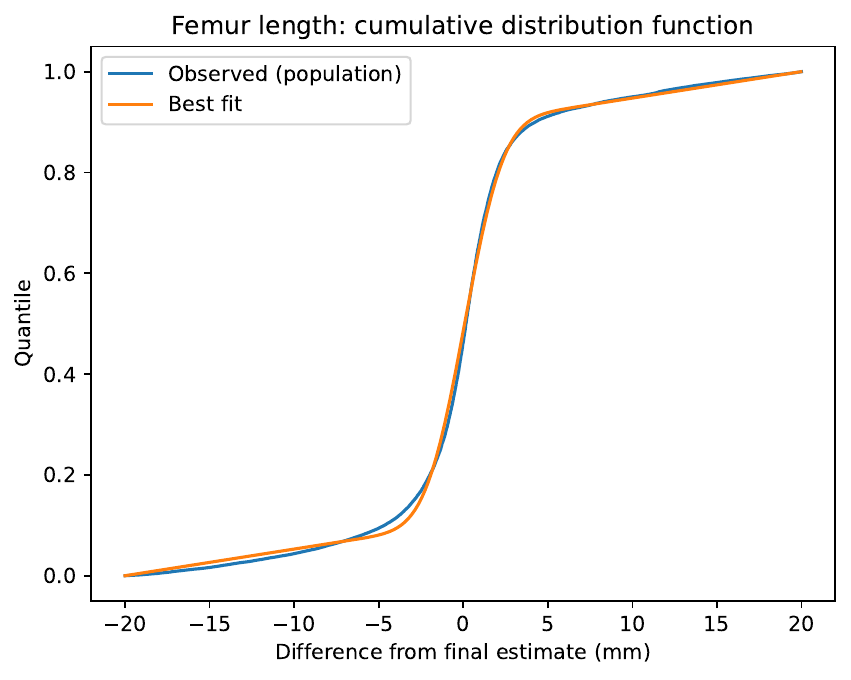}
\par\end{centering}
\begin{centering}
(a)\hspace*{0.45\columnwidth}(b)
\par\end{centering}
\caption{(a) The histogram of all femur length measurements made in individual
frames in the test set. (b) The cumulative distribution function of
all femur length measurements overlaid with the equivalent distribution
for a fitted model using a mixed Gaussian + uniform distribution.\label{fig:cumulative_dist}}
\end{figure}

Figure \ref{fig:cumulative_dist}(a) shows the distribution of all
femur length measurements across the test dataset. The values were
demeaned and then aggregated across all scans. 

Figure \ref{fig:cumulative_dist}(b) shows the cumulative distribution
function (cdf) of all measurements obtained that way, as well as the
best-fit theoretical model distribution for this data. The global
best-fit parameters for the femur length were $P_{t}=0.79$, $\mu=0$
and $\sigma=1.8$mm.

Similar curves were found for the other biometrics examined with comparable
goodness of fit.

\section{Discussion}

Our experiments show that the proposed workflow can improve the accuracy
and repeatability of biometric estimation. In single frames, our models
generally produce more variable biometric estimates than humans. However,
when estimating biometrics across an entire examination, our estimates
are in line with those made by sonographers (machine-human MSD \textasciitilde{}
human-human MSD). Paired scan data also shows that our results are
highly repeatable, with little difference across scans of the same
subject. This latter test is a direct analogue of inter-rater testing
by human observers and results were in closer agreement than for manual
measurement.

\subsection{Single-frame biometric estimation}

Section \ref{subsec:Single-frame-biometric-estimatio} compares our
model's performance in individual frames to that of a human. We generally
find good agreement with sonographer measurements. Sarris et al \citep{Sarris2012}
report inter-observer differences when two sonographers manually place
calipers on the same US frame for three biometrics. When referenced
to the corresponding human measurements, the machine measurements
in this experiment across all biometrics show more divergence than
humans do in a single frame. This appears to be random error: there
is little observed bias.

Part of this difference can be explained by failure modes of our networks.
Figure \ref{fig:Failure-frames} shows some frames for which the models
give an inaccurate biometric estimate. This typically occurs when
structures of interest are poorly visualised in the US image, or in
some edge cases such as when both femurs are visible in a frame. These
failure modes are uncommon (divergences of over 20\% as shown in Figure
\ref{fig:Failure-frames} are seen in only 1.5\% of segmentations)
but they are not mistakes that a human is likely to make.

There is significant variation in human-machine difference across
biometrics. TCD shows a very large MSD of over 10\%, while other biometrics
are significantly more closely aligned to the human measurement. Human
inter-observer variability in TCD measurements has been reported as
being between 3-5\% \citep{Chavez2003}, so the machine method appears
to show significantly greater variability.

\subsection{Whole-scan biometric estimation}

The clinical standard for biometric measurement in fetal US relies
on operator selection and annotation of a small number of individual
frames. We take a different approach: we analyse every frame in which
the biometry is visible, and use all available measurements to generate
a global estimate of the biometric. This avoids operator dependence
associated with selection of standard planes, which can be a substantial
contributor to inter-rater variability. Furthermore, it reduces the
impact of random error from caliper placement in any one plane, which
is also a significant source of variability.

The overall estimates of biometrics across a scan show better agreement
with manual measurements than those from a single image. For instance,
the TCD estimates across a scan show an average MSD with the manual
measurement of 4.37\%, compared to 10.1\% for individual frames. This
is despite the fact that the frames selected by the sonographers were
removed: aggregating measurements from a diverse and unselected range
of views of the anatomy results in a similar level of agreement with
the sonographer. This value is also comparable with human inter-observer
variability estimates found in the literature \citep{Chavez2003}.

There are also few extreme outliers, demonstrating that our system
achieves human-level performance. This is despite the fact that performance
of our CNNs on individual frames is lower than that achieved by humans:
our whole-scan estimation method can compensate for the errors introduced
by our single-frame estimation CNNs. We expect that our whole-scan
algorithm could therefore achieve superhuman performance in biometric
estimation if the single-frame estimator can be improved to human
level. The test-retest experiment we performed in Section \ref{subsec:Paired-scan-data}
provides support for this, showing the variability in biometric estimates
across scans is much lower than that between humans.

Furthermore, our method allows us to estimate credible intervals in
which a given biometric can be expected to lie. This is based on the
distribution of measurements and the number of observations, as described
in Section \ref{subsec:Whole-video-biometric-estimation}. On average,
this is smaller than the MSD by a factor of 1.4. However, the manual
measurement itself displays variability a: assuming Gaussianity, MSD
should be larger than the standard error by a factor of $\sqrt{2}\approx1.4$.
Therefore, the calculated standard error is in line with the observed
variability in our estimates. Our credible intervals therefore provide
a quantitative measure of the uncertainty in biometric estimation. 

The system discussed in this paper treats single-frame measurements
in a US scan as following a Gaussian distribution (when they are not
due to misclassifications). Section \ref{subsec:Measurement-distributions-results}
shows that this is globally an appropriate assumption, with measurements
following the expected distribution. The distribution of measurements
within each scan may be variable, depending on the operator's choice
of scanning planes, but the Bayesian estimation method ensures that
an optimal estimate is always achieved given the available data.

Another advantage of this system, when applied in real time, is that
it requires no sonographer interaction to obtain biometrics. Much
of a prenatal US scan is spent acquiring biometrics, and removing
the need for this measurement can reduce scan time by up to a third
\citep{Matthew2022}. A prospective trial of this system would be
needed to estimate and quantify a time saving. 

\subsection{Whole-scan performance and test-retest reliability}

Section \ref{subsec:Paired-scan-data} shows results of our models
on paired data from the same scan. Despite a small number of pairs,
across all biometrics our models are more repeatable and consistent
than humans. The standard deviation of the difference between measurements
of the same subject was approximately half of that between human sonographers.
Much of the remaining difference can likely be explained by the different
views of the relevant biometrics acquired in each scan determined
by sonographer technique and fetal lie. This experiment was conducted
in a sample scanned with a different US scanner and at a higher gestational
age range than the training dataset, yet the model outputs remained
robust to this domain shift.

\section{Conclusion}

This paper has demonstrated a novel method using machine learning
to estimate fetal biometrics at the 20-week scan. The proposed method
does not rely on measurements performed on individual manually-selected
standard planes, as is the norm in both manual measurements and for
many commercial machine assisted systems, but estimates biometrics
across the entire scan. This avoids the biases exhibited by humans
in plane selection and biometric measurement. We further present evidence
that estimates produced using the proposed methods may be more consistent
and repeatable than human measurements. This approach can also present
other benefits that have not been quantified in this paper: by removing
the need for sonographers to freeze the US stream and perform measurements,
the system can improve sonographer focus and reduce the time needed
to complete a FASP scan. A prospective trial of this system is in
progress to establish whether this can realise those benefits and
improve detection of fetal abnormalities.

\section*{Data availability statement}

The US scan recordings used to train the CNNs used in this paper were
collected with an ethical requirement for patient data to remain confidential.
As such, the scan data cannot be made publicly available. The raw
and processed CNN outputs used to construct our results are available
from the corresponding author on reasonable request.

\section*{Acknowledgements}

This work was supported by the Wellcome Trust/EPSRC iFIND grant (IEH
award 102431), by core funding from the Wellcome/EPSRC Centre for
Medical Engineering {[}WT203148/Z/16/Z{]} and by the National Institute
for Health Research (NIHR) Clinical Research Facility based at Guy\textquoteright s
and St Thomas\textquoteright{} NHS Foundation Trust and King\textquoteright s
College London.

\section*{Competing interests}

The authors declare the following competing interests:

All authors are co-inventors on a patent filing related to the core
methods described in this work filed by King's College London (patent
application number P333GB).

All authors are co-founders and hold equity ownership in Fraiya Ltd,
an entity that is commercialising this technology.

\bibliographystyle{ieeetr}
\bibliography{paperbib}

\newpage{}

\appendices

\section{CNN for caliper localisation\label{sec:CaliperNet}}

\subsection{Introduction}

\begin{figure}[h]
\begin{centering}
\includegraphics[width=0.48\columnwidth]{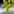}\hspace*{0.04\columnwidth}\includegraphics[width=0.48\columnwidth]{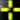}
\par\end{centering}
\caption{Zoomed examples of two calipers in the training set. These are at
different scales, with different backgrounds and levels of overlap
with other labels: it is difficult to design a machine method that
can extract both reliably. \label{fig:Caliper-shapes}}
\end{figure}

For this project, we used the sonographer's annotations taken during
their ultrasound scans as a source of data labels. These are readily
visible in the scan and trivial for a human to detect: the calipers
places by sonographers are prominent in their shape and colour. However,
there were too many calipers to feasibly extract manually: approximately
40,000 labelled images were used in this paper.

Similarly, it is difficult to use a machine algorithm to extract these
calipers automatically. Though all ultrasound scans were performed
on identical hardware, there were several software updates over the
course of the iFIND study, which changed the resolution and rendering
of calipers. These weren't always consistently rendered on the machine,
as they appeared to be interpolated and therefore displayed differently.
Furthermore, calipers often overlap with text annotations or other
calipers, making it difficult to reliably distinguish them. Figure
\ref{fig:Caliper-shapes} shows two very different caliper shapes
on different backgrounds.

We used a simple ML tool, which we called CaliperNet, to extract these
labels automatically.

\subsection{Labels and training}

\begin{figure}[h]
\centering{}\includegraphics[width=1\columnwidth]{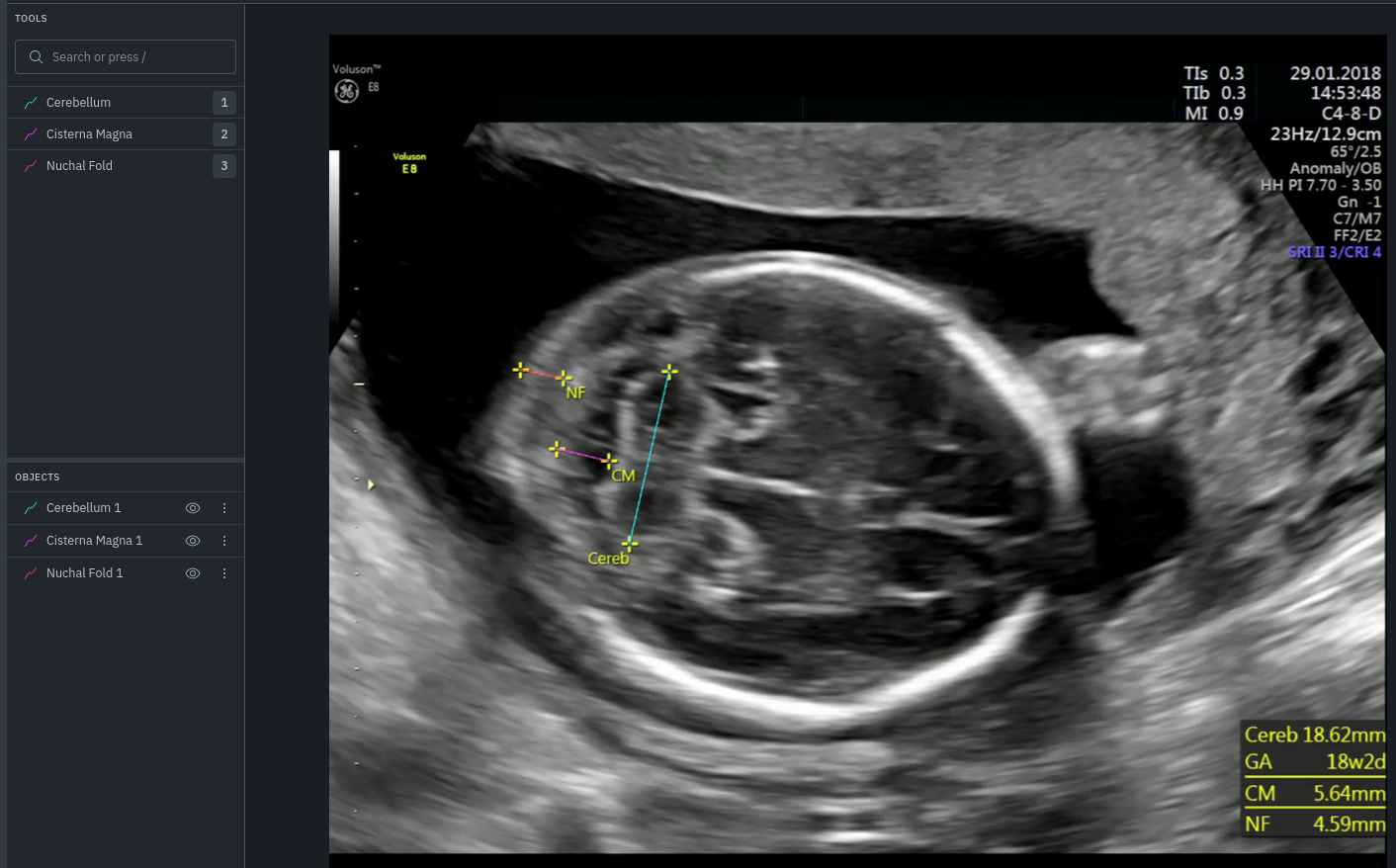}\caption{The interface of Labelbox which we used to annotate calipers, with
an example annotation of a transcerebellar image with three manual
annotations. \label{fig:Labelbox-interface}}
\end{figure}

We used a sample of 300 images per standard plane to be manually annotated
with caliper locations. We obtained these labels from a single human
annotator on Labelbox \citep{Labelbox}. Caliper locations are exact,
so we expect very little labelling noise. 

Labelbox does not have a native ellipse annotation facility, so to
trace ellipse labels we asked the annotator to pinpoint five points
along the perimeter of the ellipse: five points are sufficient to
fully constrain an ellipse's parameters. Training labels were generated
in the same way as in the main paper: for length biometrics, the endpoints
were convolved with a Gaussian kernel to generate a heatmap. For ellipse
biometrics, the outline of the ellipse was convolved with a Gaussian
kernel.

We used the same U-Net architecture as for our full biometric models
to train CaliperNet. The only difference was the retention of image
colour channels: calipers in the GE interface are coloured yellow,
while the rest of an ultrasound image is usually in greyscale (though
the operator can change the colour map), so important information
is carried by the colour channel. 

We trained three separate models: one to extract ellipse annotations
(such as HC and AC), one to extract caliper locations in the femur
image, and one to extract TCD locations in transcerebellar brain images.
The TCD model had to be trained separately as for most images in that
plane, the sonographer took three annotations and placed three sets
of calipers: the TCD, the cisterna magna (CM), and the nuchal fold
(NF)\footnote{Of these, the TCD measurement is in the FASP standard and is required
to always be measured. Nuchal fold must be measured only if it is
abnormally thick, while cisterna magna was in a previous version of
the FASP standard but is no longer required.}. Figure \ref{fig:Labelbox-interface} shows an example image with
three sets of calipers. For TCD calipers, we generated multi-channel
labels, with one channel per annotated structure. Not all images in
this standard plane had all three biometrics labelled: if calipers
were missing, the relevant channel was left blank. 

Training was performed using a 75:10:15 train:validation:test split
for images. We used the same hardware to train CaliperNet models as
the models described in the main paper. 

\subsection{Results}

\begin{table}[h]
\begin{centering}
\begin{tabular}{cccc}
\toprule 
Biometric & Bias (\%) & MSE (\%) & Dice coefficient\tabularnewline
\midrule
\midrule 
HC & -0.01\% & 0.30\% & 0.821\tabularnewline
\midrule 
AC & 0.00\% & 0.34\% & 0.819\tabularnewline
\midrule 
FL & +0.09\% & 0.66\% & 0.861\tabularnewline
\midrule 
TCD & +0.07\% & 0.95\% & 0.946\tabularnewline
\bottomrule
\end{tabular}
\par\end{centering}
\medskip{}

\caption{Results for each CaliperNet trained on one biometric.\label{tab:Calipernet-Results}}
\end{table}

\begin{figure}[h]
\begin{centering}
\includegraphics[width=0.48\columnwidth]{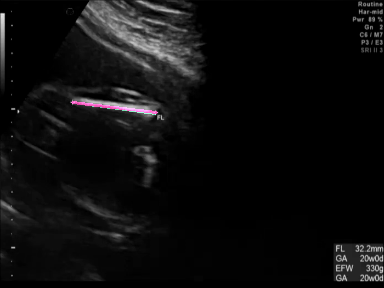}\hspace*{0.04\columnwidth}\includegraphics[width=0.48\columnwidth]{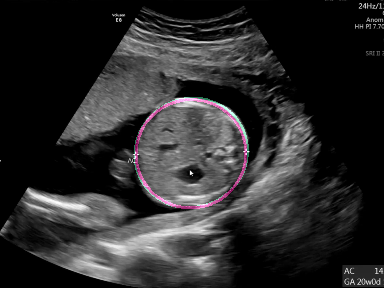}
\par\end{centering}
\centering{}(a)\hspace*{0.45\columnwidth}(b)\caption{Annotations with the largest \% errors in the test sets: green is
the ground-truth annotation and pink are the CaliperNet outputs. Shown
in (a) femur (length error of about 1\%), and (b) abdomen (about 2\%).
\label{fig:Calipernet-Failure-cases}}
\end{figure}

Table \ref{tab:Calipernet-Results} shows the performance of CaliperNet
on our test dataset across the structures of interest. There is very
little error, showing that this is a fairly straightforward task for
these networks. Often, the pixel localisations are exactly the same:
the errors are usually on the order of $\pm1$px in specific pixel
locations. The `Dice coefficient' column measures how well the network
output maps replicate the training labels. While it is high, it is
not perfect: this may be because the limited training data makes it
difficult for the CNN to learn to replicate the Gaussian kernels consistently.
Nevertheless, the regressed endpoints and ellipses remain reliable.

There were no failure cases across our test dataset. Figure \ref{fig:Calipernet-Failure-cases}
shows the images with the largest errors (-1\% and +2\%) in biometric
between the manual annotation of the calipers and the CaliperNet output.
In both cases, the difference was of 1px in endpoint localisation
- within labelling noise. This reflects the fact that caliper localisation
is a trivial task for humans, and is not a challenging task for CNNs.
Therefore, we can have confidence in the labels generated by CaliperNet
and can use them to generate training labels for our biometric CNNs. 

\newpage{}

\section{Additional performance metrics for biometric CNNs \label{sec:techreq-dump}}

\subsection{Introduction}

Two additional metrics were used to measure the performance of our
biometric CNNs and to filter for failure cases at test time. These
could give substantial indicators of failure cases, making them useful
to reject failure cases and reduce noise in our whole-video estimates.
One metric, the Dice similarity coefficient of a reconstructed heatmap,
was used to measure the quality of fit of derived ellipses and points
from the CNN's heatmap: where this was too low, the fit was deemed
of insufficient quality to accept the resulting measurement. The other
was the eccentricity of the output ellipses for head and abdominal
circumference measurements: where this was anatomically implausible,
the measurements were rejected.

\subsection{Dice similarity coefficient}

\begin{figure}[h]
\begin{centering}
\includegraphics[width=0.48\columnwidth]{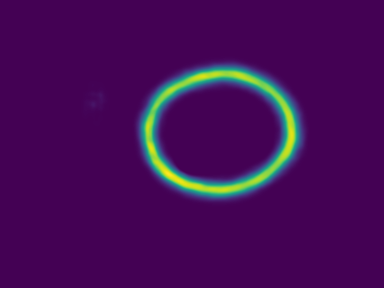}\hspace*{0.03\columnwidth}\includegraphics[width=0.48\columnwidth]{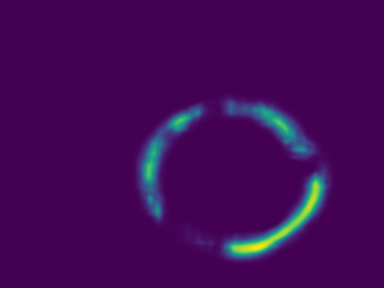}\textbf{\medskip{}
}\includegraphics[width=0.48\columnwidth]{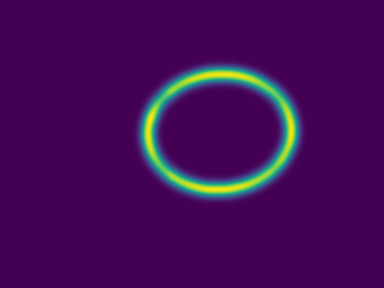}\hspace*{0.03\columnwidth}\includegraphics[width=0.48\columnwidth]{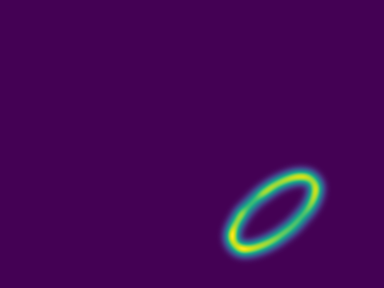}
\par\end{centering}
\begin{centering}
$\mathrm{DSC}=0.805$\hspace*{0.27\columnwidth}$\mathrm{DSC}=0.583$
\par\end{centering}
\begin{centering}
(a) \hspace*{0.45\columnwidth}(b)
\par\end{centering}
\caption{Two comparisons of the output heatmap from a CNN and the resulting
\textquoteleft reconstructed\textquoteright{} ellipse. (a) a success
case with $\mathrm{DSC}=0.805$, (b) a failure case with $\mathrm{DSC}=0.583$.
\label{fig:dice-heatmaps} }
\end{figure}

The Dice similarity coefficient (DSC) is a measure of overlap between
two sets defined as 
\[
\mathrm{DSC}(X,Y)=\frac{2\left|X\cap Y\right|}{\left|X\right|+\left|Y\right|}
\]
and is often used in image segmentation tasks to measure how closely
a segmentation follows the ground truth. 

Our biometric CNNs output heatmaps based on endpoints and ellipses,
as described in Section \ref{subsec:Biometric-measurement}. These
are then processed to return a biometric measurement: for linear measurements,
endpoints are found from local maxima of the heatmap; for ellipses,
an ellipse is fit using least-squares to the heatmap.

This can then be checked by generating a heatmap from the resulting
annotation, using the same process described in Section \ref{subsec:Biometric-measurement}.
The two heatmaps should match exactly in the case of a perfect output.
In reality, there is always some deviation, but when the reconstructed
ellipse faithfully follows the output heatmap, the resulting DSC is
high.

We found that a $\mathrm{DSC}<0.6$ between the output heatmap and
the reconstruction indicated a poor reconstruction of the CNN output.
Therefore, we discarded any proposed reconstructions that fell below
that threshold, as the output would be very noisy and inaccurate in
following anatomical boundaries.

Figure \ref{fig:dice-heatmaps} shows two example cases of reconstructed
ellipses based on an output heatmap. (a) shows a success case, where
the reconstructed ellipse closely follows the output heatmap: this
would be accepted. (b) shows a case where the output heatmap is quite
noisy and results in poor fit of the reconstructed ellipse. This has
low $\mathrm{DSC}$ and therefore is rejected. 

\subsection{Ellipse eccentricity}

\begin{figure}[h]

\begin{centering}
\includegraphics[width=0.48\columnwidth]{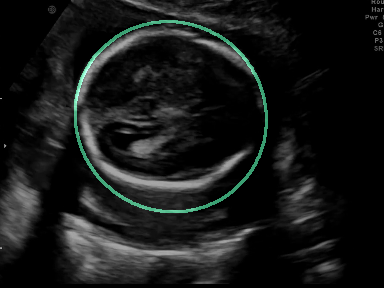}\hspace*{0.03\columnwidth}\includegraphics[width=0.48\columnwidth]{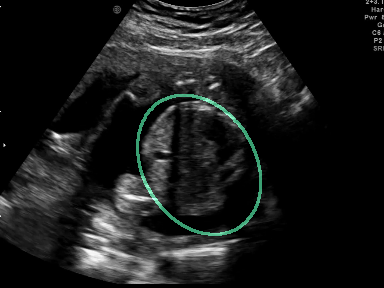}
\par\end{centering}
\begin{centering}
$e=0.242$\hspace*{0.32\columnwidth}$e=0.670$
\par\end{centering}
\begin{centering}
(a) \hspace*{0.45\columnwidth}(b)
\par\end{centering}
\caption{Biometric failure cases which violate the eccentricity constraint
in (a) the head, and (b) the abdomen. \label{fig:eccentricity-failures} }
\end{figure}

For ellipse biometrics (HC and AC), another relevant metric to consider
at test time is eccentricity. 
\[
e=\sqrt{1-\frac{a^{2}}{b^{2}}}
\]
where $a$ is the minor axis of the ellipse and $b$ is the major
axis. For the head, $a$ is the BPD measure and $b$ is the OFD (occipito-frontal
diameter). While this is a commonly used mathematical measure, in
obstetrics the more common measure for head measurements is the \textbf{cephalic
index}, defined as 
\[
CI=\frac{a}{b}\times100.
\]

In the case of an ellipse, there is a 1:1 correspondence between the
two measures, where 
\[
e=\sqrt{1-\left(\frac{CI}{100}\right)^{2}}.
\]

In the 17-22 week range of gestational age, the mean cephalic index
has been reported as 75.9, with a standard deviation of 3.7 \citep{Constantine2020}.
Therefore, an anatomically plausible range of cephalic indices is
$65-90$, which corresponds to an eccentricity range of $0.43-0.76$.
Any ellipse with an eccentricity outside of this range is likely to
be a failure mode of the biometric CNN.

In the final product using this CNN, we relaxed this constraint further
to an eccentricity range of $0.25-0.8$. This was to restrict rejections
to clear failures, allowing a large margin for biological variation.

The fetal abdomen has somewhat different morphology. It is typically
round, with a very low eccentricity. We constrained it to have $e<0.6$
for the purposes of our software - any abdominal ellipse with a higher
eccentricity was considered biologically implausible and rejected.

Figure \ref{fig:eccentricity-failures} shows two examples of drawn
ellipses which violate this eccentricity constraint. Both are clear
failures, where the ellipse does not follow anatomical boundaries.
The addition of our constraint rejects the measurements from these
ellipses as noise, even if the circumferences appear biologically
plausible.

\subsection{Plane classification confidence}

\begin{table}[h]
\begin{centering}
\begin{tabular}{cc}
\toprule 
Standard plane & Frames >95\% confidence (\%)\tabularnewline
\midrule
\midrule 
Brain-TV & 83.5\%\tabularnewline
\midrule 
Brain-CB & 82.1\%\tabularnewline
\midrule 
Abdominal & 73.2\%\tabularnewline
\midrule 
Femur & 66.4\%\tabularnewline
\bottomrule
\end{tabular}
\par\end{centering}
\centering{}\medskip{}
\caption{Proportion of frames in the four biometric standard planes which are
classified with >95\% softmax confidence.\label{tab:Frame-confidence}}
\end{table}

Section \ref{subsec:Whole-video-biometric-estimation} describes one
filtering method: only frames identified with $>95\%$ confidence
as a standard plane with a FASP biometric are processed by biometric
networks. This somewhat reduces the number of frames passed to biometric
networks and therefore the number of samples of each biometric.

Table \ref{tab:Frame-confidence} shows the proportion of frames classified
as belonging to each of the four biometric standard planes with sufficient
confidence for a measurement. In all cases, only a minority of frames
are discarded.

\end{document}